\def\year{2023}\relax
\documentclass[letterpaper]{article} 
\usepackage{aaai22}  
\usepackage{times}  
\usepackage{helvet}  
\usepackage{courier}  
\usepackage[hyphens]{url}  
\usepackage{graphicx} 
\usepackage{natbib}  
\usepackage{caption} 
\DeclareCaptionStyle{ruled}{labelfont=normalfont,labelsep=colon,strut=off} 
\usepackage{algorithm}
\usepackage{algorithmic}

\usepackage{newfloat}
\usepackage{listings}
\floatstyle{ruled}
\newfloat{listing}{tb}{lst}{}
\floatname{listing}{Listing}
\usepackage{enumerate}
\usepackage{enumitem}
\usepackage{bm}
\usepackage{amsmath}
\usepackage{amssymb}
\usepackage{amsfonts}
\usepackage{multirow}
\usepackage{tabularx}
\usepackage{mathrsfs}
\usepackage{amsthm}
\usepackage{booktabs}
\usepackage[switch]{lineno}

\title{Learned Distributed Image Compression with Multi-Scale Patch Matching in Feature Domain}
\author {
	Yujun Huang\textsuperscript{\rm 1,4},
	Bin Chen\textsuperscript{\rm 2,4}\thanks{Corresponding author.},
	Shiyu Qin\textsuperscript{\rm 2},
	Jiawei Li\textsuperscript{\rm 3},
	Yaowei Wang\textsuperscript{\rm 4},
	Tao Dai\textsuperscript{\rm 5},
	Shu-Tao Xia\textsuperscript{\rm 1,4}
}
\affiliations {
	\textsuperscript{\rm 1} Tsinghua Shenzhen International Graduate School, Tsinghua University\\
	\textsuperscript{\rm 2} Harbin Institute of Technology, Shenzhen\,\,
	\textsuperscript{\rm 3} HUAWEI Technologies Co., Ltd., Shenzhen\\
	\textsuperscript{\rm 4} Research Center of Artificial Intelligence, Peng Cheng Laboratory\,\,
	\textsuperscript{\rm 5} Shenzhen University
	huangyj20@mails.tsinghua.edu.cn, chenbin2021@hit.edu.cn, 190110427@stu.hit.edu.cn, li-jw15@tsinghua.org.cn, wangyw@pcl.ac.cn, 	daitao.edu@gmail.com, xiast@sz.tsinghua.edu.cn
}

\usepackage{bibentry}

\begin{document}
\maketitle

\begin{abstract}
	
	Beyond achieving higher compression efficiency over classical image compression codecs,  deep image compression is expected to be improved with additional side information, e.g., another image from a different perspective of the same scene. To better utilize the side information under the distributed compression scenario, the existing method~ \cite{ayzik2020deep} only implements patch matching at the image domain to solve the parallax problem caused by the difference in viewing points.  However, the patch matching at the image domain is not robust to the variance of scale, shape, and illumination caused by the different viewing angles, and can not make full use of the rich texture information of the side information image. To resolve this issue, we propose \textbf{M}ulti-\textbf{S}cale \textbf{F}eature \textbf{D}omain \textbf{P}atch \textbf{M}atching (MSFDPM) to fully utilizes side information at the decoder of the distributed image compression model. Specifically, MSFDPM consists of a side information feature extractor, a multi-scale feature domain patch matching module, and a multi-scale feature fusion network. Furthermore, we reuse inter-patch correlation from the shallow layer to accelerate the patch matching of the deep layer. 
	Finally, we find that our patch matching in a multi-scale feature domain further improves compression rate by about 20\% compared with the patch matching method at image domain~ \cite{ayzik2020deep}. 
	
\end{abstract}
\section{Introduction}
\label{sec:introduction}

\begin{figure}[t]
	\centering
	\begin{minipage}[t]{0.32\columnwidth}
		\centering
		\includegraphics[width=\linewidth]{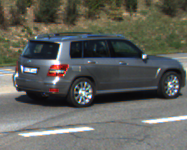}
		\centerline{SI}
		\centerline{ }
		\vspace{-4pt}
		\\ 
		
		\includegraphics[width=\linewidth]{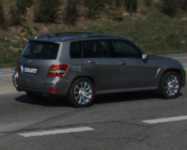} 
		\centerline{SI with weakened}
		\centerline{brightness}
		\centerline{}
		
	\end{minipage}
	\!
	\begin{minipage}[t]{0.32\columnwidth}
		\centering
		\includegraphics[width=\linewidth]{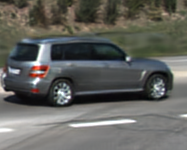} 
		\centerline{\footnotesize{(a) 0.0384 bpp}}
		\centerline{\footnotesize{0.9058 MS-SSIM}}
		\vspace{-4pt}
		\\ 
		\includegraphics[width=\linewidth]{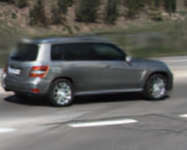} 
		\centerline{\footnotesize{(a') 0.0384 bpp}}
		\centerline{\footnotesize{0.8975 MS-SSIM}}
		
	\end{minipage}%
	\,
	\begin{minipage}[t]{0.32\columnwidth}
		\centering
		\includegraphics[width=\linewidth]{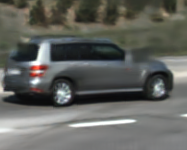} 
		\centerline{\footnotesize{(a) 0.0360 bpp}}
		\centerline{\footnotesize{0.8829 MS-SSIM}}
		\vspace{-4pt}
		\\ 
		\includegraphics[width=\linewidth]{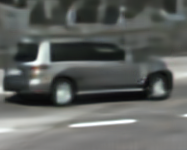} 
		\centerline{\footnotesize{(b') 0.0360 bpp}}
		\centerline{\footnotesize{0.8709 MS-SSIM}}
	\end{minipage}
	\centering
	\vspace{0pt}
	\caption{Visual comparisons of decoded main images with different compression methods when the brightness of the side information (SI) image is normal or weakened. (a) and (b) are examples of path matching in the multi-scale feature domain (proposed) and patch matching in the image domain, respectively, when the brightness is normal. (a') and (b') are the corresponding examples when the brightness of side information image is $60\%$ of the original image. The visual perception of (a) is similar to (a'). However, (b') loses some color and texture details compared to (b). } 
	\label{fig:feature_vs_image_visualization_brightness_10_and_06}
\end{figure}


Distributed image compression is less studied than the commonly known single image compression, as the inherent correlation among images taken from different viewpoints is hard to capture. With the development of the stereo camera, camera array, self-driving system, and multiple UAV/monitor camera systems during the past decade, high-volume multi-view images become the primary data source in large-scale digital communication. For example, it has been reported that self-driving cars can capture 1G of data per second \cite{mearian2013self}. Since multi-view images are usually captured from different angles of the same scene at the same time, there exist overlapping fields with high correlation among these images.
In order to improve the compression efficiency for such distributed compression scenario, the inherent correlation must be utilized to remove redundancy in data. This inspires the development of distributed compression with side information, where 
the sensors encode their data, independently, and the decoders recover the data with the help of side information from another source only available at the decoder. The Distributed Source Coding (DSC) theorem \cite{slepian1973noiseless, 1055356, wyner1976rate} reveals that the same compression rate can be asymptotically achieved by using side information (SI) only at the decoding end as when available at both encoder and decoder. However, there is still a performance gap between these two coding schemes in practice.

Recently, with the success of deep learning, deep image compression \cite{a54241fecb3e4d9a8a25b36a984b410c, 47602, mentzer2018conditional} has made great improvement over classical codecs like JPEG. In a deep image compression model, a variational autoencoder is adopted as a basic architecture for end-to-end optimized encoding and decoding. For a given input image, the multiple nonlinear layers extract its latent feature representation. And the entropy of the quantized feature layer is estimated by an entropy model. Then the quantized feature is input into a symmetric decoder to reconstruct the image. By optimizing the rate-distortion (R-D) loss over the large-scale training set, deep image compression achieves state-of-the-art compression ratios.


In recent years, some learning-based distributed image coding methods have been proposed \cite{diao2020drasic, ayzik2020deep, mital2022neural}. Ayzik et al. \cite{ayzik2020deep} proposed to align the side information image with the main image by patch matching in the image domain. However, operating only in the image domain can not make full use of the abundant texture details in side information, and is not robust to the variants of illumination and scale caused by different viewing points. As shown in Fig. \ref{fig:feature_vs_image_visualization_brightness_10_and_06}, the gain of the side information in the image domain achieves a slight improvement when the brightness of the side image is weakened. By contrast, our proposed multi-scale feature domain patch matching method still makes good use of the correlative texture information within the side information image. This significantly demonstrates the robustness and efficiency of the multi-scale feature domain patch matching for side information utilization.



We introduce a \textbf{M}ulti-\textbf{S}cale \textbf{F}eature \textbf{D}omain \textbf{P}atch \textbf{M}atching (MSFDPM) method to better utilize the side information image at the decoder of the deep image compression model.
Similarity matching in the multi-scale feature domain can fully explore the correlative texture information in the side information image and is robust to variants of illumination and scale. At the same time, the complementary scale features can effectively recover the decoded main image from fine-grained texture to coarse-grained structures simultaneously. Specifically, MSFDPM consists of three modules: (1) A feature extractor network to extract multi-scale side information features from side information image; (2) Patch matching module to obtain aligned side information features, in which the inter-patch correlation of the largest scale feature layer is reused to later feature layers to accelerate the decoding process; (3) Feature fusion network to concatenate the decoded  multi-scale features from the main image with that of the side information image to obtain high-quality reconstruction. 

To summarize, we make the following contributions.
\setlist{nolistsep}
\begin{itemize}[leftmargin=1em]
	\item We propose a new paradigm of	multi-scale feature domain patch matching method for deep distributed image compression. 
	
	\item We validate that the patch matching in the multi-scale feature domain is more robust to variants of illumination and scale than patch matching in the image domain.
	
	\item We provide comprehensive experiments to show that our method outperforms the state-of-the-art distributed image compression methods. 
\end{itemize}
\section{Related Work}
\label{sec:related_work}
\subsubsection{Deep Image Compression}
\label{subsec:DIC}
Deep image compression has attracted more and more attention in recent years. Thanks to the powerful representation capability of autoencoder, flexible entropy model, and end-to-end optimization architecture, deep image compression methods have outperformed reigning image compression standards. Ballé \textit{et al.} first proposed an end-to-end autoencoder structure based on rate-distortion optimization for image compression \cite{a54241fecb3e4d9a8a25b36a984b410c}. Ballé \textit{et al.} further proposed a hyperprior model to capture the spatial correlation of latent feature map and transfer the captured information to the decoder as side information \cite{47602}. Minnen \textit{et al.} proposed an autoregressive prior and a hierarchical prior to obtain the correlation between neighboring components in the latent feature map \cite{minnen2018joint}. Cheng \textit{et al.} proposed to use the discretized mixture Gaussian model to exactly model the distribution of latent feature map and introduced the attention modules into the autoencoder \cite{9156817}. Xie \textit{et al.} proposed to use an Invertible Encoding Network instead of the autoencoder structure to mitigate information loss \cite{xie2021enhanced}. Kim \textit{et al.} proposed a novel entropy model to focus on long-range correlations in hidden layer features through an attention mechanism \cite{Kim_2022_CVPR}.

\begin{figure*}[t]
	\centering
	\includegraphics[width=\textwidth]{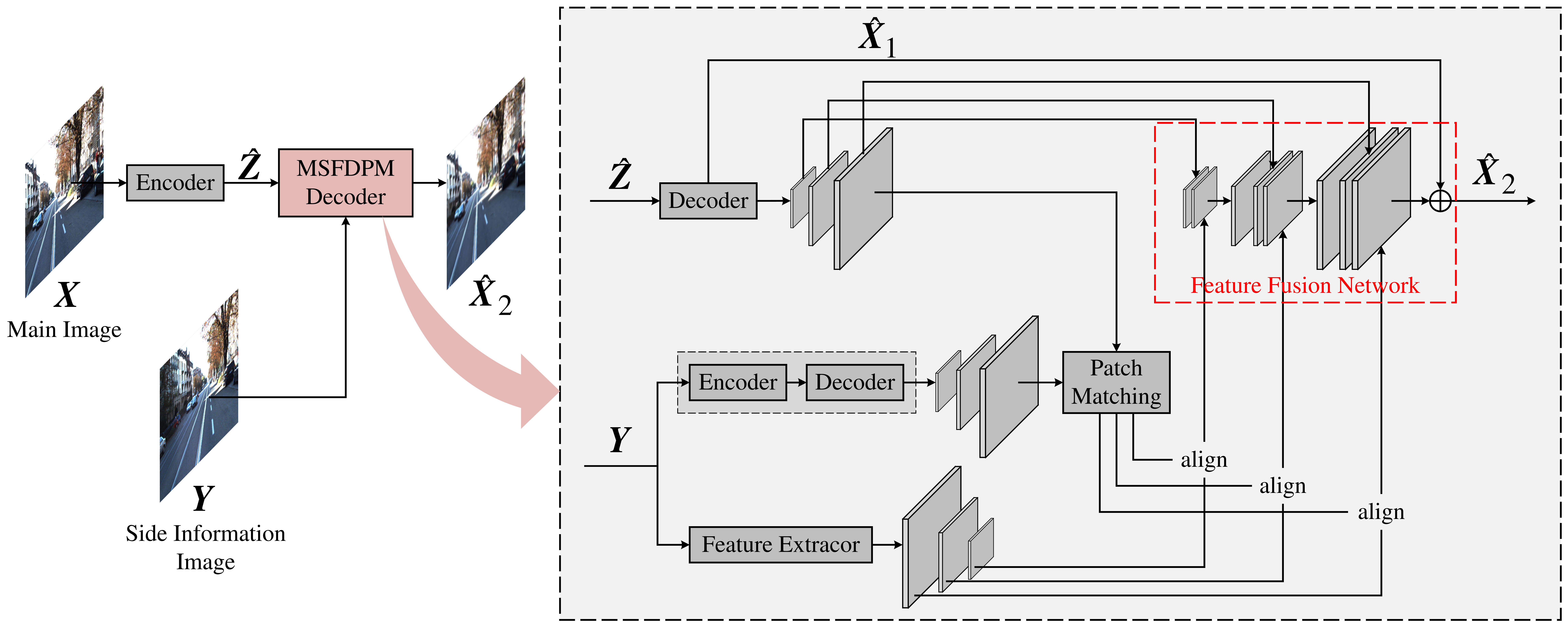}
	\caption{The proposed MSFDPM architecture for side information based decoding of deep image compression }
	\label{fig:framework}
\end{figure*}

\subsubsection{Learned Distributed Source Coding}
\label{subsec:LDSC}
Recently, some papers use deep learning to implement distributed source coding. Ding \textit{et al.} proposed to use recurrent autoencoder for distributed image coding, but without using side information images when decoding main image \cite{diao2020drasic}. It only uses decoder model to learn the correlation.  Ayzik \textit{et al.} proposed that the side information image could be aligned with the main image by patch matching method when there is a large overlapping field between them, and the aligned side information image could be used to obtain better main image reconstruction \cite{ayzik2020deep}. However, the information available in image domain is limited and not robust enough. Mital \textit{et al.} proposed to extract the common information from the side information image through a network and input the common information and the latent feature map of main image into a decoder network to obtain an enhanced reconstruction \cite{mital2022neural}. However, the proposed method does not align the side information image with main image, which will lead to the degraded reconstruction quality when the parallax is large.

\section{The Proposed Method}
\label{sec:method}
The main idea of our proposed MSFDPM decoder is to transmit both high-level semantic information and low-level texture information to the main image based on multi-scale feature domain patch matching, so that useful information can be screened out for reconstruction. Specifically, our method first uses a feature extractor to obtain multi-scale side information features, aligns the features with the main features by patch matching, and then fuses the aligned side information features with main features to get a better reconstruction. The proposed method is shown in Fig. \ref{fig:framework}. 

\subsection{Multi-Scale Features of Main Image and Side Information Image}
\label{subssec:msf_of_mm_and_sim}

Before patch matching, we need the model to obtain multi-scale features of the main image and side information image. First, the main image $\textbf{\textit{X}}$ is fed into the encoder of the single-image compression model to obtain latent feature map $\hat{\textbf{\textit{Z}}}$. Because single-image decoding is an upsampling process, the decoder could take the latent feature map $\hat{\textbf{\textit{Z}}}$ and output multi-scale decoded main features $\left\{\left.\textbf{\textit{F}}_{\hat{\textbf{\textit{X}}}}^{\textit{h}}\in\mathbb{R}^{\frac{H}{2^h}\times \frac{W}{2^h}\times C} \right| h=1,2,3,4\right\}$ and the first stage decoded main image $\hat{\textbf{\textit{X}}}_1\in \mathbb{R}^{H\times W\times C}$, where $H$ and $W$ are the height and width of the main image. We use smaller superscripts to represent larger scale features, which facilitates the following description of patch matching. With the same single-image autoencoder, the side information image $\textbf{Y}$ can be used to produce multi-scale decoded side information features $\left\{\left.\textbf{\textit{F}}_{\hat{\textbf{\textit{Y}}}}^{\textit{h}}\in\mathbb{R}^{\frac{H}{2^h}\times \frac{W}{2^h}\times C} \right| h=1,2,3,4\right\}$. Because the decoded main features and the decoded side information features are produced by the same autoencoder, they have a similar property, which will improve matching accuracy. However, the decoded side information features lose part of the side information because of the quantization noise, lossless side information features are needed to improve the main image reconstruction. A trainable feature extractor is proposed to take the side information image and produce the multi-scale lossless side information features $\left\{\left.\textbf{\textit{F}}_{\textbf{\textit{Y}}}^{\textit{h}}\in\mathbb{R}^{\frac{H}{2^h}\times \frac{W}{2^h}\times C} \right| h=1,2,3,4\right\}$. Our feature extractor uses the encoder structure in \cite{9156817}.

\subsection{Multi-Scale Feature Domain Patch Matching}
\label{subsec:msfdpm}

Because of the disparity problem caused by different viewing points, the side information image needs to be aligned with the main image to effectively use the side information. In addition, compared to the image domain, which would be more affected by variance of illumination and color caused by different viewing points, the feature domain will lay more emphasis on the texture and structure to be migrated. Moreover, because it is time-consuming to calculate the inter-patch correlation on all feature layers, we choose to calculate only this correlation on the first feature layer, which can be reused by other layers. For the convenience of description, we define the set of small patches sampled by a $B\times B$ window with stride $s$ on a $H\times W\times C$ feature map $T$ as:
\begin{eqnarray}
	& S^{T,B,s}=\{p_{i,j}^{T, B, s} \mid i=0, \cdots, I, j=0, \cdots, J \}\\
	&\text{where}\,\, I\triangleq\left\lfloor \frac{W-B}{s} \right\rfloor,
	J\triangleq \left\lfloor \frac{H-B}{s} \right\rfloor \nonumber 
\end{eqnarray}
According to this definition, we can obtain the patch set of the first decoded main feature layer $S^{\textbf{\textit{F}}_{\hat{\textbf{\textit{X}}}}^{1},B,B}$, the first decoded side information feature layer $S^{\textbf{\textit{F}}_{\hat{\textbf{\textit{Y}}}}^{1},B,1}$, and multi-scale lossless side information $\left\{\left.S^{\textbf{\textit{F}}_{\textbf{\textit{Y}}}^{h},\frac{B}{2^{h-1}},1}\right|h=1,2,3,4\right\}$. It is worth noting that the patches of the decoded main feature are not overlapping and cover all the regions of the feature map, while the patch sampling of the side information feature is fine-grained because we need to match the most similar patch on the side information feature map. 
We use the pearson correlation coefficient to measure the inter-patch correlation between patches of the first decoded main feature layer and the first decoded side information feature layer:
\begin{equation}
	r_{(i,j),(k,l)} = {\rm Pr}(
	p_{i,j}^{\textbf{\textit{F}}_{\hat{\textbf{\textit{X}}}}^{1}, B, B}, 
	p_{k,l}^{\textbf{\textit{F}}_{\hat{\textbf{\textit{Y}}}}^{1}, B, 1})
	*m_{(i,j),(k,l)},
\end{equation}
where $m_{(i,j),(k,l)}$ is the Gaussian mask proposed in \cite{ayzik2020deep}, which is the prior to choose adjacent patches with higher probability.
The computationally expensive inner product operation in the pearson correlation can be efficiently calculated by a series of convolutions, where the convolution kernel is each patch of the decoded main feature, with the decoded side information feature as input:
\begin{equation}
	p_{i,j}^{\textbf{\textit{F}}_{\hat{\textbf{\textit{X}}}}^{1}, B, B}
	* \textbf{\textit{F}}_{\hat{\textbf{\textit{Y}}}}^{1}, 
\end{equation}
where $*$ denotes the convolution operation. Based on the pearson correlation coefficient, we can find the location of the side information feature patch that are most similar to each patch in the main feature:
\begin{equation}
	(k_{(i,j)}^{*}, l_{(i,j)}^{*}) = \mathop{\arg\max}\limits_{(k,l)} r_{(i,j),(k,l)}.
\end{equation}
Then we can get the first aligned side information feature layer $\textbf{\textit{F}}_{\textbf{\textit{Y}};a}^{1}$ by putting the most similar lossless side information feature patch into the corresponding position in the main feature:
\begin{equation}
	p_{i,j}^{\textbf{\textit{F}}_{\textbf{\textit{Y}};a}^{1}, B, B}=
	p_{k_{(i,j)}^{*}, l_{(i,j)}^{*}}^{\textbf{\textit{F}}_{\textbf{\textit{Y}}}^{1}, B, 1}.
	\label{eq:align_the_first_lossless_side_information_feature_layer}
\end{equation}

\subsubsection{Reusing First Feature Layer Inter-Patch Correlation}
\label{subsubsec:rfflic}
We reduce the computational complexity and GPU memory footprint of patch matching by reusing the above-mentioned first feature layer inter-patch correlation.
First, we correspond the patches of the second to the fourth feature layer to the patches of the same position in the first feature layer. A toy example is shown in Appendix A. We find that for the decoded main features the position subscripts of patches corresponding to the same position in different feature layers are the same. For the decoded side information features, the position subscript of the patch in the previous layer is twice that of the corresponding patch in the later layer. So we can use the correlation between the corresponding patches of the first feature layer to represent the inter-patch correlation of the following layer:
\begin{equation}
	\begin{split}	
		R(
		p_{i,j}^{\textbf{\textit{F}}_{\hat{\textbf{\textit{X}}}}^{h}, \frac{B}{2^{h-1}}, \frac{B}{2^{h-1}}},
		p_{k,l}^{\textbf{\textit{F}}_{\hat{\textbf{\textit{Y}}}}^{h},
			\frac{B}{2^{h-1}}, 1})
		&= r_{(i,j),(2^{h-1}k, 2^{h-1}l)},\\
		&\qquad \qquad h=2,3,4.
	\end{split}	
\end{equation}
Then we can get the second to the fourth aligned side information feature layers $\left\{\textbf{\textit{F}}_{\textbf{\textit{Y}};a}^{2}, \textbf{\textit{F}}_{\textbf{\textit{Y}};a}^{3}, \textbf{\textit{F}}_{\textbf{\textit{Y}};a}^{4}\right\}$ by using the method similar to Eq.  \ref{eq:align_the_first_lossless_side_information_feature_layer}.

\begin{figure}[t]
	\centering
	\includegraphics[width=\columnwidth]{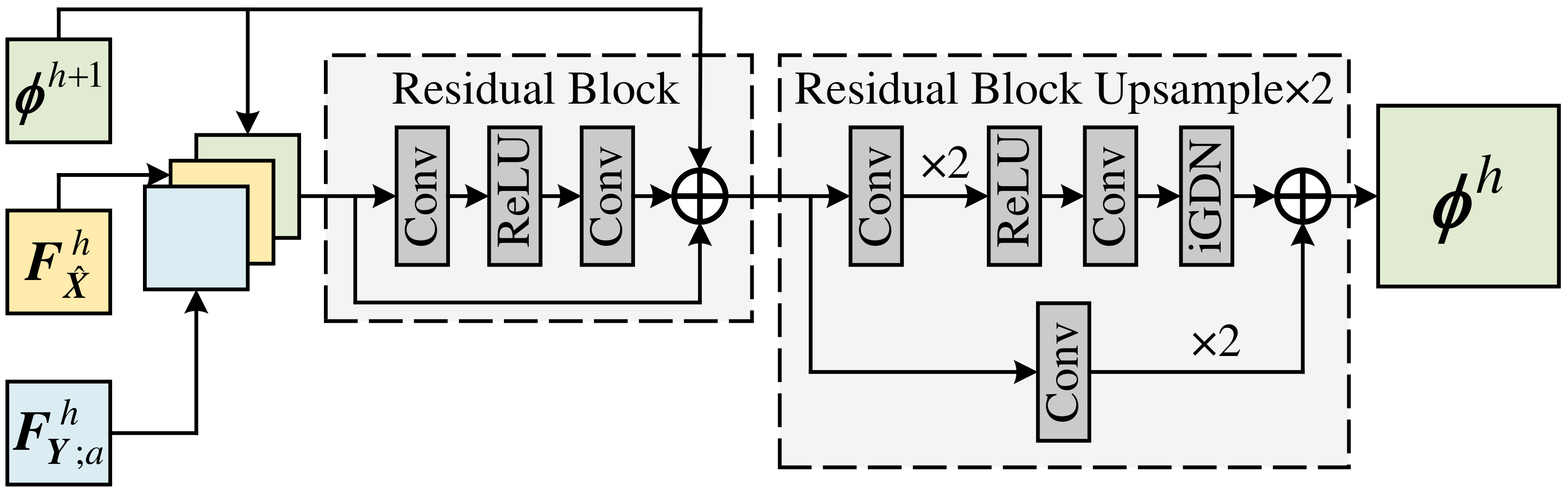}
	\caption{One iteration of feature fusion network. 
	}
	\label{fig:feature_fusion_network}
\end{figure}

\subsection{Feature Fusion Network}
\label{subsec:ffn}

Feature fusion network is designed to iteratively fuse aligned side information features and decoded main features from small scale to large scale, as illustrated in the red box in Fig. \ref{fig:framework}. The effectiveness of feature fusion at multiple scales is demonstrated in ablation experiments. Specifically, in one iteration of the feature fusion network, the aligned side information feature and the decoded main feature at scale $h$, as well as the output feature of the previous iteration (if any) are concatenated and input into two residual blocks to obtain the fused feature at scale $h$:
\begin{equation}
	\begin{split}	
		\bm{\phi}^{4} &= {\rm Res}^4_2\left({\rm Res}^4_1\left(
		\textbf{\textit{F}}_{\hat{\textbf{\textit{X}}}}^{4},
		\textbf{\textit{F}}_{\textbf{\textit{Y}};a}^{4}
		\right)\right),
		\\
		\bm{\phi}^{h} &= {\rm Res}^h_2\left({\rm Res}^h_1\left(
		\textbf{\textit{F}}_{\hat{\textbf{\textit{X}}}}^{h},
		\textbf{\textit{F}}_{\textbf{\textit{Y}};a}^{h},
		\bm{\phi}^{h+1}
		\right)
		+\bm{\phi}^{h+1}
		\right),\\
		&\ \ \ \ \ \ \ \ \qquad \qquad \qquad \qquad \qquad h=1,2,3. 
	\end{split}	
\end{equation}
The architectures of the residual blocks refer to the decoder of single-image compression model \cite{9156817}. Fig. \ref{fig:feature_fusion_network} illustrates in detail one iteration of the feature fusion network. Finally, the second stage decoded main image is obtained by adding the latest output of the feature fusion network to the first stage decoded main image:
\begin{equation}
	\hat{\textbf{\textit{X}}}_2 = \bm{\phi}^1 + \hat{\textbf{\textit{X}}}_1
\end{equation}

\subsection{Loss Function}
\label{subsec:lf}

Lossy compression is a joint optimization problem of compression rate and distortion. For the proposed model, the loss function consists of the entropy of the latent feature map $\hat{\textbf{\textit{Z}}}$, the distortion of the first stage decoded main image and the second stage decoded main image:
\begin{equation}
	\mathcal{L} = H(\hat{\textbf{\textit{Z}}}) + \lambda((1-\alpha)d(
	\textbf{\textit{X}},
	\hat{\textbf{\textit{X}}}_1)
	+\alpha d(
	\textbf{\textit{X}},
	\hat{\textbf{\textit{X}}}_2)
	)),
\end{equation}
where $\lambda$ is the weight that controls trade-off between the compression rate and the distortions. $\alpha$ is the weight that controls the trade-off between the two distortions. 

\begin{figure*}[t]
	\centering
	\includegraphics[width=0.8\linewidth]{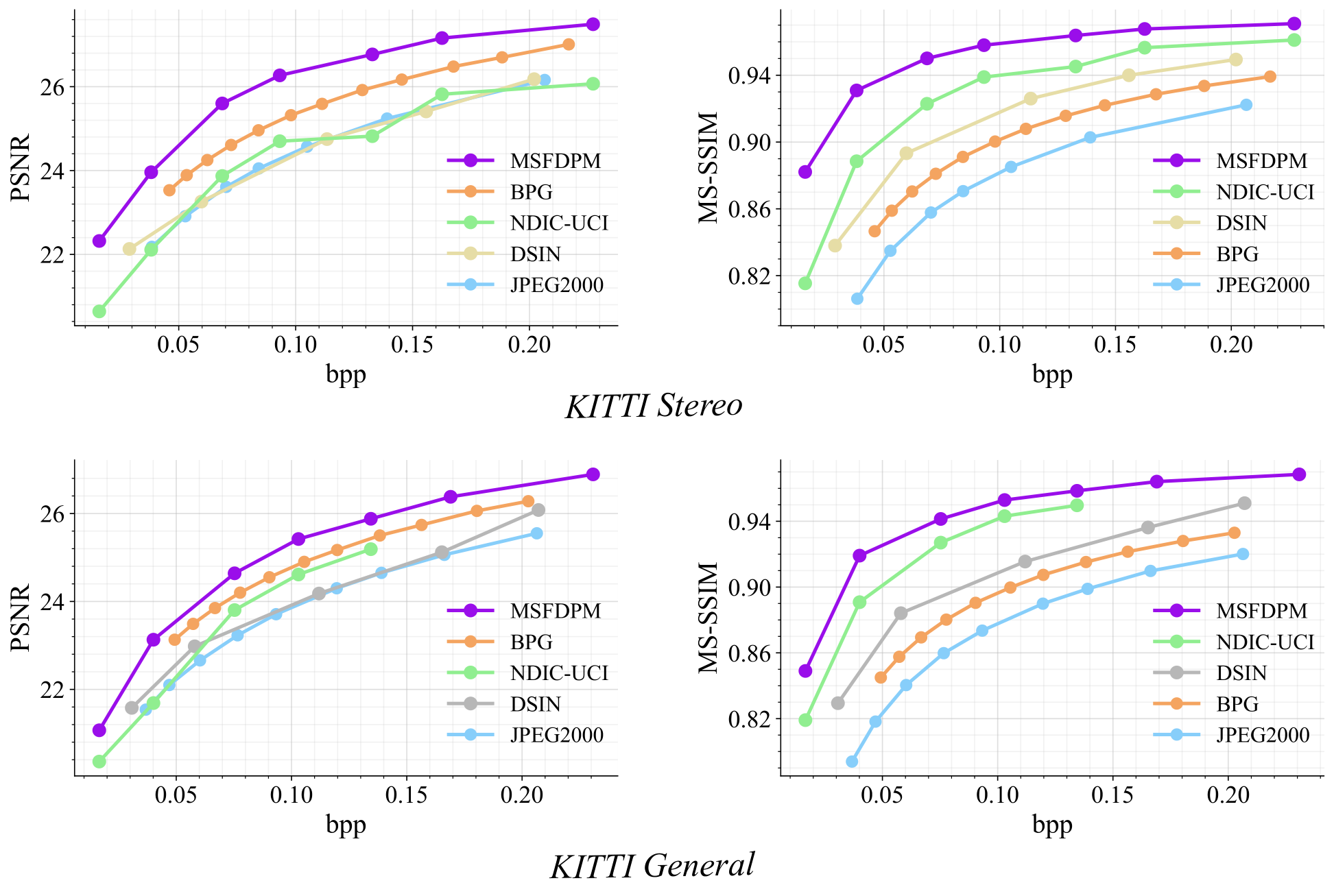}
	\caption{The rate-distortion performance comparison of different methods. 
	}
	\label{fig:si_msf}
\end{figure*}

\section{Experiments and Results}
\label{sec:experiments}

\subsection{Experimental Setup}
\label{subsec:experiemntal_setup}

\subsubsection{Datasets}
\label{subsubsec:datasets}

We conduct experiments on two datasets: \emph{KITTI Stereo} and \emph{KITTI General} proposed in \cite{ayzik2020deep}. \emph{KITTI Stereo} contains 1578 training pairs and 790 test pairs, which are paired stereo images from the KITTI Stereo 2012 \cite{Geiger2012CVPR} and KITTI Stereo 2015\cite{Menze2015CVPR} dataset. \emph{KITTI General} has 174936 training pairs and 3609 test pairs. The pairs of data are different not only in viewing points but also in time steps.

\subsubsection{Evaluation Metrics}
\label{subsubsec:metrics}

We use bits per pixel (bpp) to measure the compression ratio. Both peak signal-to-noise ratio (PSNR) and multi-scale structural similarity (MS-SSIM) \cite{wang2003multiscale} are used to measure image quality/signal distortion. Moreover, we use BD-Rate, negative numbers to indicate the percentage of average bit savings at the same image quality across the rate-distortion curve compared with some chosen baselines. Specifically, BD-Rate$_\text{P}$ and BD-Rate$_\text{M}$ indicate the performance gain under the same PSNR and MS-SSIM, respectively.

\subsubsection{Baselines}
\label{subsubsec:baselines}

We compare the MSFDPM with 4 baselines. Two classical image compression baselines: JPEG 2000\cite{10.1117/1.1469618} and BPG \cite{BPG-web}. And two learned distributed image compression baselines: DSIN \cite{ayzik2020deep} and NDIC-UCI \cite{mital2022neural}. DSIN and NDIC-UCI are briefly introduced in related work. 

\subsubsection{Implementation Details}
\label{subsubsec:implementation_details}

The proposed MSFDPM is implemented with PyTorch~\cite{PyTorch} and the experiments are conducted on four Intel(R) Xeon(R) E5-2698 v4 CPUs and eight NVIDIA Tesla V100 GPUs. We first train a single-image compression baseline \cite{9156817}. Then we train the full model with cropped $320\times 960$ image pair, where the parameters of the autoencoder are initialized with the pretrained baseline. The number of epochs on \emph{KITTI Stereo} is 10 and the number of epochs on \emph{KITTI General} is 1. We used a batch size of 1 and the Adam optimizer \cite{kingma2015adam} with $1\cdot 10^{-4}$ learning rate. 
Other hyper-parameters are listed as follows: 
(\textbf{i}) The number of features, $C=128$.
(\textbf{ii}) The patch size, $B=16$.
(\textbf{iii}) The weight for rate-distortion trade-off, $\lambda \in \left\{0.005, 0.01, 0.02, 0.035, 0.05, 0.07, 0.1\right\}$.
(\textbf{iv}) The weight for two stages of distortions, $\alpha$ is equal to $0$ when training the autoencoder baseline and $1$ when training the full model.

\begin{figure*}[thbp]
	\centering
	\begin{minipage}[t]{0.24\linewidth}
		\centering
		\includegraphics[width=\linewidth]{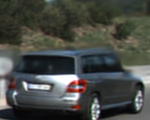}
		\centerline{\footnotesize{0.0391 bpp/0.9292 MS-SSIM}}
		\vspace{0pt}
		\centerline{\tiny{ }} 
		\centerline{\large{ }}
		\vspace{-1pt}
		\\ 
		
		\includegraphics[width=\linewidth]{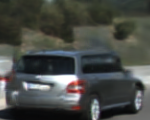} 
		\centerline{\footnotesize{0.0367 bpp/0.8908 MS-SSIM}} 
		\centerline{\tiny{ }} 
		
	\end{minipage}
	\!
	\begin{minipage}[t]{0.24\linewidth}
		\centering
		\includegraphics[width=\linewidth]{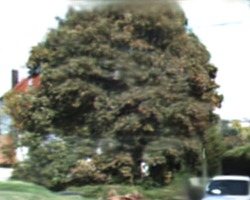} 
		\centerline{\footnotesize{0.0365 bpp/0.9298 MS-SSIM}}  
		\vspace{0pt}
		\centerline{\tiny{ }} 
		\centerline{\large{\qquad\qquad\qquad\qquad\qquad\ \ \ \ \ \ MSFDPM}}
		\vspace{-1pt}
		\\ 
		
		\includegraphics[width=\linewidth]{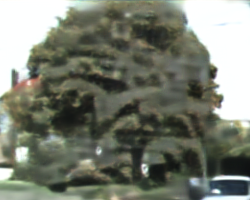} 
		\centerline{\footnotesize{0.0331 bpp/0.9053 MS-SSIM}} 
		\centerline{\tiny{ }} 
		\centerline{\large{\qquad\qquad\qquad\qquad\qquad\ \ \ \ \ \ \ \ MSFDPM$_\text{img}$}} 
	\end{minipage}%
	\,
	\begin{minipage}[t]{0.24\linewidth}
		\centering
		\includegraphics[width=\linewidth]{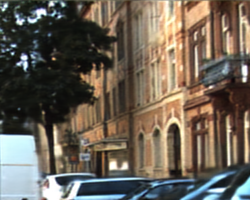} 
		\centerline{\footnotesize{0.1094 bpp/0.9541 MS-SSIM}}  
		\vspace{0pt}
		\centerline{\tiny{ }} 
		\centerline{\large{ }}
		\vspace{-1pt}
		\\ 
		
		\includegraphics[width=\linewidth]{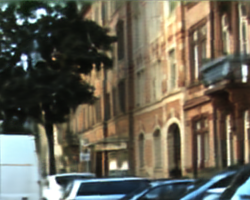} 
		\centerline{\footnotesize{0.1047 bpp/0.9433 MS-SSIM}} 
		\centerline{\tiny{ }}  
	\end{minipage}
	\!
	\begin{minipage}[t]{0.24\linewidth}
		\centering
		\includegraphics[width=\linewidth]{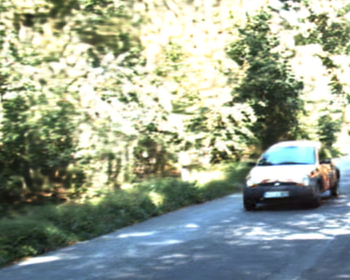} 
		\centerline{\footnotesize{0.1362 bpp/0.9019 MS-SSIM}}  
		\vspace{0pt}
		\centerline{\tiny{ }} 
		\centerline{\large{ }}
		\vspace{-1pt}
		\\ 
		\includegraphics[width=\linewidth]{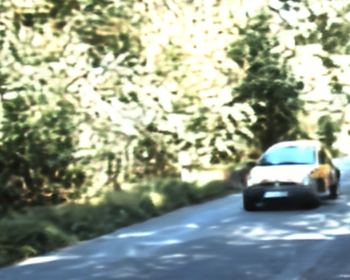} 
		\centerline{\footnotesize{0.1303 bpp/0.8900 MS-SSIM}} 
		\centerline{\tiny{ }}  
	\end{minipage}
	\centering
	\vspace{0pt}
	\caption{Visual comparisons of MSFDPM and MSFDPM$_\text{img}$. } 
	\label{fig:feature_vs_image_visualization}
\end{figure*}

\subsection{Results and Analysis}
\label{subsec:analysis}

\subsubsection{Results}
\label{subsec:results}

We report the rate-distortion results in Fig. \ref{fig:si_msf}, which shows that our MSFDPM outperforms all comparison baselines across different compression ratios.
The increase in compression is significant. For example, in the KITTI Stereo dataset, when the MS-SSIM is $0.90$, the bpp of MSFDPM is about $0.025$, that of NDIC-UCI is about $0.048$, and that of DSIN is about $0.07$, that of BPG is about $0.098$, and that of JPEG 2000 is about $0.134$. Compared with NDIC-UCI, DSIN, BPG, and JPEG 2000, MSFDPM has a compression rate increase of $1.92$ times, $2.80$ times and $3.92$ times, $5.36$ times, respectively.

\begin{table}[t]
	\caption{The BD-Rate results of different MSFDPM variants. Baseline is single-image compression model \cite{9156817}.}
	\centering
	\resizebox{\columnwidth}{!}{
		\setlength{\tabcolsep}{.5em}{
			\begin{tabular}{lllll}
				\toprule
				\multirow{2}{*}{Variant} & \multicolumn{2}{c}{\emph{KITTI Stereo} }                             & \multicolumn{2}{c}{\emph{KITTI General}}                          \\
				\cmidrule(l){2-3} \cmidrule(l){4-5} 
				& BD-Rate$_\text{P}$ & BD-Rate$_\text{M}$ & BD-Rate$_\text{P}$ & BD-Rate$_\text{M}$\\
				\midrule
				
				MSFDPM & \textbf{-49.26\%}  & \textbf{-51.11\%}  & \textbf{-40.37\%}  & \textbf{-40.89\%}  \\
				
				MSFDPM$_{\text{img}}$  & -27.81\% & -24.54\%  & -18.72\% & -13.53\% \\
				
				MSFDPM$_{\text{f1}}$ & -41.29\% & -42.02\% & -31.15\% & -29.92\% \\
				
				MSFDPM$_\text{f2}$ & -45.99\% & -49.13\% & -38.77\% & -38.50\% \\
				
				MSFDPM$_\text{f3}$ & -25.43\% & -33.25\% & -32.72\% & -35.12\% \\
				
				MSFDPM$_{\text{f4}}$ & -24.80\% & -33.22\% & -24.41\% & -27.89\% \\
				
				MSFDPM$_{ \text{w/o r}}$ & -47.01\% & -49.89\% & -40.06\% & -40.41\% \\
				\bottomrule
	\end{tabular}}}
	\label{tab:variants}
\end{table}

\subsubsection{Ablation Experiments}
\label{subsubsec:ablation_experiment}

To explore the effect of multi-scale feature patch matching and inter-patch correlation reusing, we design 3 variants of MSFDPM:
(\textbf{i}) MSFDPM$_\text{img}$ replaces multi-scale feature domain patch matching with image domain patch matching. The aligned side information image and the first stage decoded main image are input into a network containing four residual blocks to obtain fine reconstruction.
(\textbf{ii}) MSFDPM$_\text{f$h$}$, $h=1,2,3,4$, only fuses the $h$-th aligned side information feature layer in the feature fusion network.
(\textbf{iii}) MSFDPM$_\text{w/o r}$ performs block matching at each feature layer without reusing the inter-patch correlation of the first feature layer.  
\begin{itemize}
	\item \textbf{Multi-Scale Feature Domain Patch Matching v.s. Image Domain Patch Matching.} MSFDPM outperforms MSFDPM$_\text{img}$ by BD-Rate$_\text{P}$ of $-21.45\%$ and BD-Rate$_\text{M}$ of $-26.57\%$ on \emph{KITTI Stereo} and BD-Rate$_\text{P}$ of $-31.65\%$ and BD-Rate$_\text{M}$ of $-27.36\%$ on \emph{KITTI General}. This implies that the multi-scale feature domain can extract more texture or structure information to help obtain higher quality reconstruction compared to the image domain.  Fig. \ref{fig:feature_vs_image_visualization} presents some visual comparisons of MSFDPM and MSFDPM$_\text{img}$. Benefiting from the multi-scale feature patch matching, MSFDPM can provide finer texture structure and richer color details than MSFDPM$_\text{img}$. 
	\item \textbf{Multi-Scale Feature Fusion v.s. Single-Scale Feature Fusion.}
	MSFDPM outperforms MSFDPM$_\text{f1}$, MSFDPM$_\text{f2}$, MSFDPM$_\text{f3}$ and MSFDPM$_\text{f4}$ on both BD-Rate$_\text{P}$ and BD-Rate$_\text{M}$. 
	This indicates that multi-scale feature fusion can improve the quality of reconstruction by using complementary fine-grained texture information to coarse-grained structure information.  
	Specifically, MSFDPM outperforms MSFDPM$_\text{f1}$, MSFDPM$_\text{f2}$, MSFDPM$_\text{f3}$, MSFDPM$_\text{f4}$ by an average BD-Rate$_\text{P}$ of $-8.59\%$, $-2.43\%$, $-15.74\%$, $-20.21\%$  and an average BD-Rate$_\text{M}$ of $-10.03\%$, $-2.18\%$, $-11.82\%$, $-15.45\%$. The BD-rate performance of MSFDPM$_\text{f2}$ is the best among all the single-scale feature fusion models. This may be because the second feature layer has a better trade-off between high-frequency detailed texture and low-frequency structural information. The performance of MSFDPM$_\text{f3}$ and MSFDPM$_\text{f4}$ is worse than that of MSFDPM$_\text{f1}$ and MSFDPM$_\text{f2}$, which indicates that texture information is more useful to improve image quality than structure information.
\end{itemize}

\begin{figure}[t]
	\centering
	\includegraphics[width=\columnwidth]{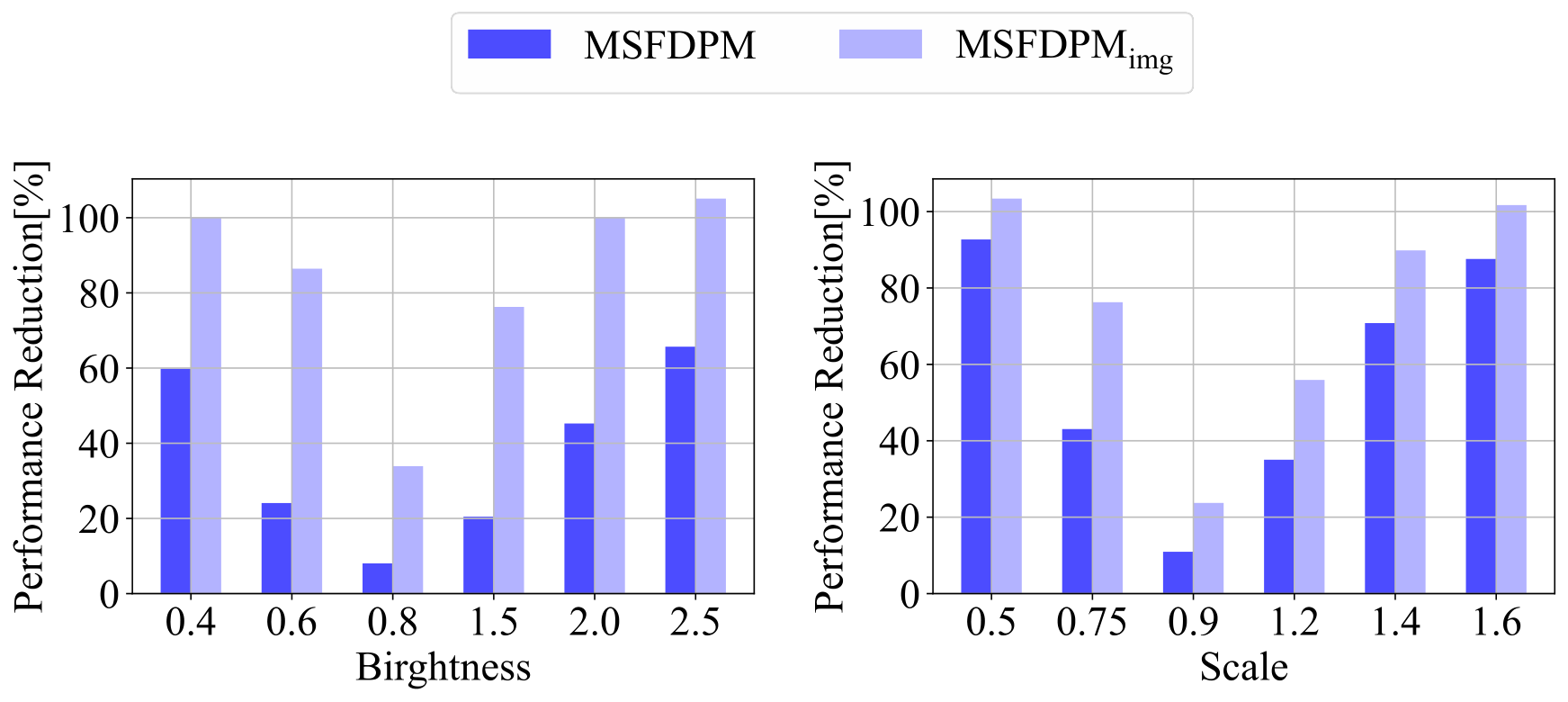}
	\caption{The performance reduction of MS-SSIM of MSFDPM and MSFDPMimg as side information image brightness and scale change ($\lambda=0.035$). 
	}
	\label{fig:robustness_experiments}
\end{figure}

\begin{figure*}[t]
	\centering
	\begin{minipage}[t]{0.05\linewidth}
		\rotatebox{90}{\LARGE  SI with scale 1} \vspace{17pt}\\
		
		\rotatebox{90}{\LARGE  SI with scale 0.75}
	\end{minipage}
	\begin{minipage}[t]{0.25\linewidth}
		\centering
		\includegraphics[width=\linewidth]{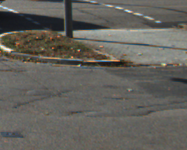}
		\centerline{SI}
		\centerline{}
		\vspace{-4pt}
		\\ 
		
		\includegraphics[width=\linewidth]{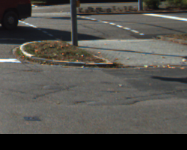} 
		\centerline{SI}
		\centerline{}
		
	\end{minipage}
	\!
	\begin{minipage}[t]{0.25\linewidth}
		\centering
		\includegraphics[width=\linewidth]{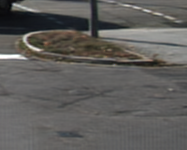} 
		\centerline{MSFDPM}
		\centerline{\fontsize{9pt}{\baselineskip}\selectfont0.0368/0.9498}
		\vspace{-4pt}
		\\ 
		\includegraphics[width=\linewidth]{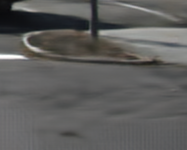} 
		\centerline{MSFDPM}
		\centerline{\fontsize{9pt}{\baselineskip}\selectfont0.0368/0.9113/83.33\%}
		
	\end{minipage}%
	\,
	\begin{minipage}[t]{0.25\linewidth}
		\centering
		\includegraphics[width=\linewidth]{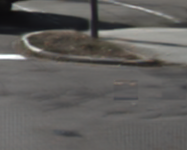} 
		\centerline{MSFDPM$_\text{img}$}
		\centerline{\fontsize{9pt}{\baselineskip}\selectfont0.0356/0.9196}
		\vspace{-4pt}
		\\ 
		\includegraphics[width=\linewidth]{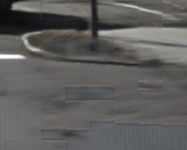} 
		\centerline{MSFDPM$_\text{img}$}
		\centerline{\fontsize{9pt}{\baselineskip}\selectfont0.0356/0.8995/125.62\%}
	\end{minipage}
	\centering
	\vspace{0pt}
	\caption{Visual comparisons of MSFDPM and MSFDPM$_\text{img}$ with scale 1 and 0.75. Evaluation Metric is denoted by bpp/MS-SSIM/PR. } 
	\label{fig:feature_vs_image_visualization_scale_10_and_075}
\end{figure*}

\subsubsection{Robustness Experiments}
\label{subsubsec:robustness_experiment}

In order to explore the robustness of the multi-scale feature domain and image domain to illumination and scale change, we adjuest the brightness and scale of the side information image during inference. We set the brightness and scale of the original side information image are 1, and proportionally change the brightness and scale of the original image. To evaluate the performance reduction (PR), we define it as 
\begin{equation}
	\text{PR}=1-\frac{\text{performance improvement after change}}{\text{performance improvement before change}}  
\end{equation}
As shown in Fig.~6, when the brightness is $0.8$, the numerical MS-SSIM improvement of the model using side information is $0.015$ compared with the model without using side information. When the brightness is $1$, the numerical MS-SSIM improvement of the model using side information is $0.02$. Then the performance reduction at $0.8$ brightness is $1-\frac{0.015}{0.02}=25\%$. 

MSFDPM outperforms MSFDPM$_\text{img}$ by an average of performance reduction of $-46.39\%$ and $-18.45\%$ with brightness change and scale change, respectively.
Besides, it can be seen that the performance reduction of MSFDPM is consistently lower than that of MSFDPM$_\text{img}$ across different brightness and scales. Therefore, we can achieve the conclusion that the multi-scale feature domain is more robust to the variants of illumination and scale than the image domain. Besides, the performance reduction of the multi-scale feature domain is much lower during brightness change. For example, when the brightness is 0.75, the performance reduction of MSFDPM$_\text{img}$ is $86.44\%$ and the performance reduction of MSFDPM is only $24.09\%$. This demonstrates that the feature domain can mitigate the brightness variation and emphasize texture and structure. 

Fig. \ref{fig:feature_vs_image_visualization_scale_10_and_075} shows the visualization results when the scale of the side information image changes. It can be seen that when the scale is $0.75$, the example of MSFDPM$_\text{img}$ shows a clustering block effect while the example of MSFDPM does not. This may be because the change of scale causes the corresponding patches cannot be accurately matched in the image domain. Appendix B shows more visualization results. 

\begin{table}[t]
	\caption{Decoding speed, generation speed of Gaussian mask, and GPU memory usage of MSFDPM and MSFDPM$_{ \text{w/o r}}$.}
	
	\centering
	\resizebox{\columnwidth}{!}{
		\setlength{\tabcolsep}{.5em}{
			\begin{tabular}{lllll}
				\toprule
				Variant & Decoding & Guassian Masks & GPU Memory \\
				\midrule
				
				MSFDPM & \textbf{0.5404s} & \textbf{22.86s}  & \textbf{9908Mb} \\
				
				MSFDPM$_{ \text{w/o r}}$ & 0.5740s & 25.30s & 10791Mb \\
				\bottomrule
	\end{tabular}}}
	\label{tab:efficiency_analysis}
\end{table}

\subsubsection{Efficiency Analysis}
\label{subsubsec:efficiency_analysis}

To verify the efficiency of reusing the first feature layer inter-patch correlation, we compare the decoding speed, Gaussian mask generation speed, and GPU memory usage of MSFDPM and MSFDPM$_{ \text{w/o r}}$, and the results are shown in Table \ref{tab:efficiency_analysis}. The decoding speed and Gaussian mask generation speed of MSFDPM are $6\%$ and $11\%$ faster than MSFDPM$_{ \text{w/o r}}$. And MSFDPM has $8\%$ less GPU memory usage than MSFDPM$_{ \text{w/o r}}$. Besides, as shown in Table \ref{tab:variants}, the BD-rate of MSFDPM is slightly lower than that of MSFDPM$_{ \text{w/o r}}$. This may be because the later decoded feature layers are closer to the quantization and are more likely inaccurately match patches due to the quantization noise. 

\section{Conclusions}
\label{sec:conclusion}

In this paper, we propose multi-scale feature domain patch matching (MSFDPM) for distributed image compression. Compared with the previous patch matching in the image domain, MSFDPM is more robust to the variants of illumination and scale caused by different viewing points, and the feature domain contains richer texture information in the side information image. In addition, we reuse the first feature layer inter-patch correlation to improve the decoding efficiency. On the experimental side, we demonstrate the superiority of MSFDPM over state-of-the-art baseline, while being more robust to variants of illumination and scale.

\section{Acknowledge}
This work was done when Yujun Huang was an internship at Harbin Institute of Technology, Shenzhen. This work is supported in part by the National Natural Science Foundation of China under grant 62171248, the PCNL KEY project (PCL2021A07), the Guangdong Basic and Applied Basic Research Foundation under grant 2021A1515110066, and the GXWD 20220811172936001, and Shenzhen Science and Technology Program under Grant JCYJ20220818101012025.

\bibliography{dsc_reference}

\begin{thebibliography}{21}
\providecommand{\natexlab}[1]{#1}

\bibitem[{Ayzik and Avidan(2020)}]{ayzik2020deep}
Ayzik, S.; and Avidan, S. 2020.
\newblock Deep image compression using decoder side information.
\newblock In \emph{European Conference on Computer Vision}, 699--714. Springer.

\bibitem[{Ball{\'e}, Laparra, and
  Simoncelli(2017)}]{a54241fecb3e4d9a8a25b36a984b410c}
Ball{\'e}, J.; Laparra, V.; and Simoncelli, E. 2017.
\newblock End-to-end optimized image compression.
\newblock 5th International Conference on Learning Representations, ICLR 2017 ;
  Conference date: 24-04-2017 Through 26-04-2017.

\bibitem[{Ballé et~al.(2018)Ballé, Minnen, Singh, Hwang, and
  Johnston}]{47602}
Ballé, J.; Minnen, D.; Singh, S.; Hwang, S.~J.; and Johnston, N. 2018.
\newblock Variational Image Compression with a Scale Hyperprior.
\newblock In \emph{6th Int. Conf. on Learning Representations ({ICLR})}.

\bibitem[{Bellard(2014)}]{BPG-web}
Bellard, F. 2014.
\newblock BPG Image format.
\newblock \url{https://bellard.org/bpg/}.

\bibitem[{Cheng et~al.(2020)Cheng, Sun, Takeuchi, and Katto}]{9156817}
Cheng, Z.; Sun, H.; Takeuchi, M.; and Katto, J. 2020.
\newblock Learned Image Compression With Discretized Gaussian Mixture
  Likelihoods and Attention Modules.
\newblock In \emph{2020 IEEE/CVF Conference on Computer Vision and Pattern
  Recognition (CVPR)}, 7936--7945.

\bibitem[{Cover(1975)}]{1055356}
Cover, T. 1975.
\newblock A proof of the data compression theorem of Slepian and Wolf for
  ergodic sources (Corresp.).
\newblock \emph{IEEE Transactions on Information Theory}, 21(2): 226--228.

\bibitem[{Diao, Ding, and Tarokh(2020)}]{diao2020drasic}
Diao, E.; Ding, J.; and Tarokh, V. 2020.
\newblock Drasic: Distributed recurrent autoencoder for scalable image
  compression.
\newblock In \emph{2020 Data Compression Conference (DCC)}, 3--12. IEEE.

\bibitem[{Geiger, Lenz, and Urtasun(2012)}]{Geiger2012CVPR}
Geiger, A.; Lenz, P.; and Urtasun, R. 2012.
\newblock Are we ready for Autonomous Driving? The KITTI Vision Benchmark
  Suite.
\newblock In \emph{Conference on Computer Vision and Pattern Recognition
  (CVPR)}.

\bibitem[{Kim, Heo, and Lee(2022)}]{Kim_2022_CVPR}
Kim, J.-H.; Heo, B.; and Lee, J.-S. 2022.
\newblock Joint Global and Local Hierarchical Priors for Learned Image
  Compression.
\newblock In \emph{Proceedings of the IEEE/CVF Conference on Computer Vision
  and Pattern Recognition (CVPR)}, 5992--6001.

\bibitem[{Kingma and Ba(2015)}]{kingma2015adam}
Kingma, D.~P.; and Ba, J. 2015.
\newblock Adam: A Method for Stochastic Optimization.
\newblock In \emph{ICLR (Poster)}.

\bibitem[{Mearian(2013)}]{mearian2013self}
Mearian, L. 2013.
\newblock Self-driving cars could create 1GB of data a second.
\newblock \emph{Computerworld}, 23.

\bibitem[{Mentzer et~al.(2018)Mentzer, Agustsson, Tschannen, Timofte, and
  Van~Gool}]{mentzer2018conditional}
Mentzer, F.; Agustsson, E.; Tschannen, M.; Timofte, R.; and Van~Gool, L. 2018.
\newblock Conditional probability models for deep image compression.
\newblock In \emph{Proceedings of the IEEE Conference on Computer Vision and
  Pattern Recognition}, 4394--4402.

\bibitem[{Menze and Geiger(2015)}]{Menze2015CVPR}
Menze, M.; and Geiger, A. 2015.
\newblock Object Scene Flow for Autonomous Vehicles.
\newblock In \emph{Conference on Computer Vision and Pattern Recognition
  (CVPR)}.

\bibitem[{Minnen, Ball{\'e}, and Toderici(2018)}]{minnen2018joint}
Minnen, D.; Ball{\'e}, J.; and Toderici, G.~D. 2018.
\newblock Joint autoregressive and hierarchical priors for learned image
  compression.
\newblock \emph{Advances in neural information processing systems}, 31.

\bibitem[{Mital et~al.(2022)Mital, {\"O}zy{\i}lkan, Garjani, and
  G{\"u}nd{\"u}z}]{mital2022neural}
Mital, N.; {\"O}zy{\i}lkan, E.; Garjani, A.; and G{\"u}nd{\"u}z, D. 2022.
\newblock Neural distributed image compression using common information.
\newblock In \emph{2022 Data Compression Conference (DCC)}, 182--191. IEEE.

\bibitem[{Paszke et~al.(2019)Paszke, Gross, Massa, Lerer, Bradbury, Chanan,
  Killeen, Lin, Gimelshein, Antiga et~al.}]{PyTorch}
Paszke, A.; Gross, S.; Massa, F.; Lerer, A.; Bradbury, J.; Chanan, G.; Killeen,
  T.; Lin, Z.; Gimelshein, N.; Antiga, L.; et~al. 2019.
\newblock PyTorch: An Imperative Style, High-Performance Deep Learning Library.
\newblock \emph{Advances in Neural Information Processing Systems}, 32:
  8026--8037.

\bibitem[{Rabbani(2002)}]{10.1117/1.1469618}
Rabbani, M. 2002.
\newblock {JPEG2000: Image Compression Fundamentals, Standards and Practice}.
\newblock \emph{Journal of Electronic Imaging}, 11(2): 286.

\bibitem[{Slepian and Wolf(1973)}]{slepian1973noiseless}
Slepian, D.; and Wolf, J. 1973.
\newblock Noiseless coding of correlated information sources.
\newblock \emph{IEEE Transactions on information Theory}, 19(4): 471--480.

\bibitem[{Wang, Simoncelli, and Bovik(2003)}]{wang2003multiscale}
Wang, Z.; Simoncelli, E.~P.; and Bovik, A.~C. 2003.
\newblock Multiscale structural similarity for image quality assessment.
\newblock In \emph{The Thrity-Seventh Asilomar Conference on Signals, Systems
  \& Computers, 2003}, volume~2, 1398--1402. Ieee.

\bibitem[{Wyner and Ziv(1976)}]{wyner1976rate}
Wyner, A.; and Ziv, J. 1976.
\newblock The rate-distortion function for source coding with side information
  at the decoder.
\newblock \emph{IEEE Transactions on information Theory}, 22(1): 1--10.

\bibitem[{Xie, Cheng, and Chen(2021)}]{xie2021enhanced}
Xie, Y.; Cheng, K.~L.; and Chen, Q. 2021.
\newblock Enhanced invertible encoding for learned image compression.
\newblock In \emph{Proceedings of the 29th ACM International Conference on
  Multimedia}, 162--170.

\end{thebibliography}
\newpage



\end{document}